%% file: main.tex
\documentclass[10pt,twocolumn,letterpaper]{article}

\usepackage{iccv}              

\input{preamble}

\definecolor{iccvblue}{rgb}{0.21,0.49,0.74}
\usepackage[pagebackref,breaklinks,colorlinks,allcolors=iccvblue]{hyperref}

\makeatletter
\def\customsymbol#1{
    \ifcase\number\value{#1}
        \or *
        \or $\dagger$
        \or 1
    \else\@ctrerr
    \fi
}
\makeatother

\def\paperID{5637} 
\def\confName{ICCV}
\def\confYear{2025}

\newcommand{\methodname}{\textsc{Neurons}}

\title{\methodname{}: Emulating the Human Visual Cortex Improves Fidelity and Interpretability in fMRI-to-Video Reconstruction}

\author{Haonan Wang\footnotemark[1], \  Qixiang Zhang\footnotemark[1], \ Lehan Wang, Xuanqi Huang, Xiaomeng Li\textsuperscript{$\dagger$}\\
The Hong Kong University of Science and Technology\\
{\tt\small \{hwanggr, qzhangcq, lwangdk, xhuangdp\}@connect.ust.hk, eexmli@ust.hk}
}

\begin{document}
\maketitle

\setcounter{footnote}{1}
\renewcommand{\thefootnote}{\customsymbol{footnote}}
\footnotetext[1]{Equal contribution.}
\setcounter{footnote}{2}
\renewcommand{\thefootnote}{\customsymbol{footnote}}
\footnotetext[2]{Corresponding author: eexmli@ust.hk.} 

\input{sec/0_abstract}    
\input{sec/1_intro}
\input{sec/2_related_works}

\input{sec/3_method}

\input{sec/4_exp}

\input{sec/5_conclusion}
{
\small
\bibliographystyle{ieeenat_fullname}
\bibliography{main, mybib}
}

\input{sec/X_suppl}

\end{document}

%% file: preamble.tex
\PassOptionsToPackage{table}{xcolor}
\usepackage{multirow}
\usepackage{color}
\usepackage{booktabs}
\usepackage{caption}
\usepackage{colortbl}
\usepackage[ruled]{algorithm2e}
\usepackage{pifont}
\usepackage{graphicx}
\usepackage{marvosym}
\usepackage{url}
\usepackage{animate}

\definecolor{commentcolor}{RGB}{110,154,155}   

%% file: sec/0_abstract.tex
\begin{abstract}
Decoding visual stimuli from neural activity is essential for understanding the human brain. While fMRI methods have successfully reconstructed static images, fMRI-to-video reconstruction faces challenges due to the need for capturing spatiotemporal dynamics like motion and scene transitions. Recent approaches have improved semantic and perceptual alignment but struggle to integrate coarse fMRI data with detailed visual features. Inspired by the hierarchical organization of the visual system, we propose NEURONS, a novel framework that decouples learning into four correlated sub-tasks: key object segmentation, concept recognition, scene description, and blurry video reconstruction. This approach simulates the visual cortex's functional specialization, allowing the model to capture diverse video content. In the inference stage, NEURONS generates robust conditioning signals for a pre-trained text-to-video diffusion model to reconstruct the videos. Extensive experiments demonstrate that NEURONS outperforms state-of-the-art baselines, achieving solid improvements in video consistency (26.6\%) and semantic-level accuracy (19.1\%). 
Notably, NEURONS shows a strong functional correlation with the visual cortex, highlighting its potential for brain-computer interfaces and clinical applications.
Code and model weights are available at: \href{https://github.com/xmed-lab/NEURONS}{\textit{\texttt{https://github.com/xmed-lab/NEURONS}}}.
\end{abstract}

%% file: sec/1_intro.tex
\section{Introduction}

\begin{figure}
\centering
    \includegraphics[width=1\linewidth]{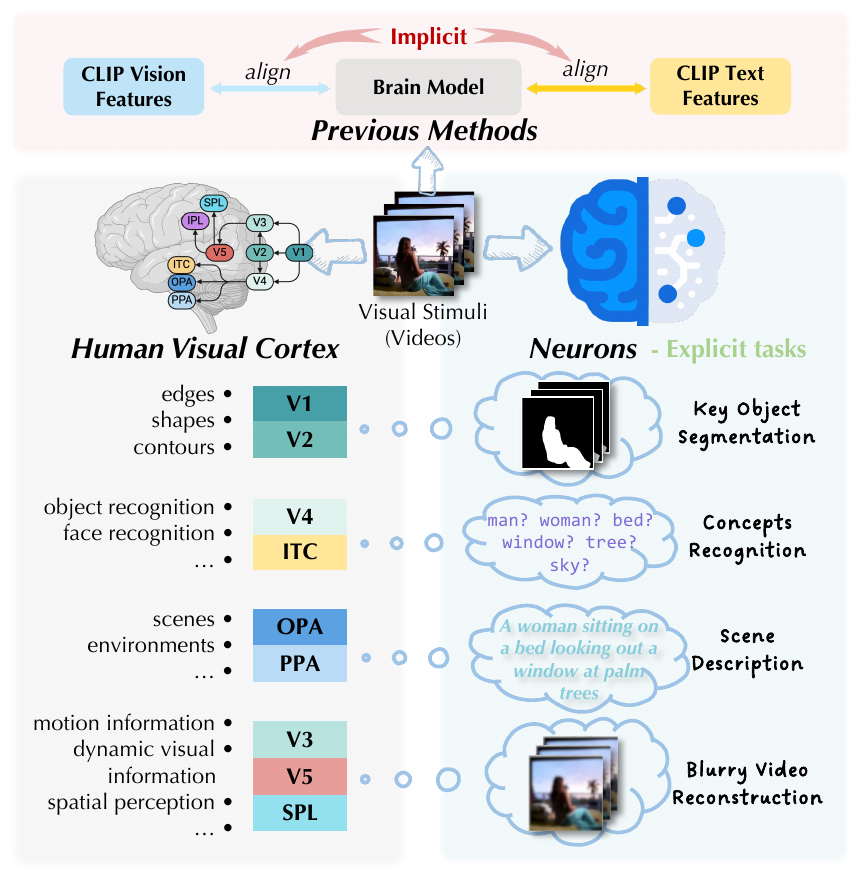}
    \caption{Previous methods~\cite{chen2024cinematic,gong2025neuroclips} implicitly align the brain model with the highly coupled CLIP vision and text hidden features. 
    Research has demonstrated that distinct regions of the human visual cortex are responsible for processing different types of visual information.
    Drawing inspiration from the human visual cortex, where distinct regions specialize in processing different aspects of visual information, we decouple the fMRI-to-video reconstruction task into four corresponding sub-tasks, thereby capturing the full spectrum of visual information in video stimuli.}
    \label{fig:abstract}
\end{figure}

Decoding visual stimuli from neural activity offers a critical pathway to unraveling the intricate mechanisms of the human brain.
Current approaches have demonstrated remarkable success in leveraging non-invasive functional magnetic resonance imaging (fMRI) data to reconstruct static images (fMRI-to-image) ~\cite{scotti2024mindeye,scotti2024mindeye2,nishimoto2011reconstructing,gong2024mindtuner} by capitalizing on advanced deep learning architectures such as CLIP~\cite{radford2021clip} and Stable Diffusion~\cite{rombach2022ldm}.
These frameworks bridge neuroscience and artificial intelligence, enabling data-driven exploration of the cerebral cortex’s functions.

However, extending these advances to fMRI-to-video reconstruction remains a formidable challenge. Unlike static images, video decoding requires capturing the spatiotemporal dynamics of continuous visual streams—including object motions, scene transitions, and temporal coherence—which existing models struggle to represent holistically. 
Early studies~\cite{nishimoto2011reconstructing,wen2018neural,wang2022reconstructing,kupershmidt2022penny} faced challenges in achieving satisfactory reconstruction performance as they struggled to extract precise semantics from pre-trained diffusion models. MinD-Video~\cite{chen2024cinematic} addresses semantic issues by conditioning the diffusion model on visual fMRI features, but it lacks low-level visual detailing, diverging from the brain’s visual system and limiting perception of continuous details. NeuroClips~\cite{gong2025neuroclips} advances the field by introducing a semantics reconstructor for accurate video keyframe reconstruction and a perception reconstructor for blurry video reconstruction. However, the main reconstruction process relies on aligning the brain model with CLIP features \textit{implicitly}, \textit{i.e.}, aligning in the hidden feature space, which is not robust. The reason is that the CLIP hidden feature space is highly semantic but the fMRI voxels are encoded with multi-granularity information (Fig.~\ref{fig:abstract}).

Emulating the functional architecture of the visual cortex offers a promising strategy to improve the decoding of brain activity. Prior research has demonstrated that distinct regions within the visual cortex are specialized for processing different levels of visual information granularity~\cite{vc_milner2017two,vc_huff2018neuroanatomy,vc_liu2024organization,vc_wang2024large} (see Fig.~\ref{fig:abstract}). Inspired by the hierarchical organization of the human visual system, we design the learning framework as a composition of multiple sub-tasks, each reflecting specific perceptual roles. For example, early visual areas such as V1 and V2 are involved in encoding low-level visual features—such as edges and shapes—which parallels the function of segmentation modules in computer vision.
However, segmenting every object in a video is impractical due to the inherent limitations of human memory in retaining detailed video information. 
Since individuals predominantly focus on \textit{key objects} and salient features within scenes rather than processing all finer details~\cite{keyobj_judd2009learning},
we introduce the \textit{key object segmentation task}, which trains the brain model to initially learn the shapes and contours of key objects. Subsequently, we progressively incorporate more complex semantic concepts and motion information through the \textit{concept recognition task}, which mimics the object and face recognition processes in areas V4 and ITC, compensating for the semantic loss incurred by focusing solely on key objects. Then, the \textit{scene description task} enables the brain model to identify correlations between concepts and generate coherent scene descriptions. Finally, the \textit{blurry video reconstruction task} integrates motion and color distribution information. These four tasks are learned using a progressive strategy that mirrors the hierarchical structure of the visual cortex.

To construct these tasks, we leverage off-the-shelf vision-language foundation models to generate key object segmentation masks, class labels, and scene descriptions. Building on this, we propose a novel framework named \methodname{}, which simulates the functional behaviors of neurons in the human visual cortex. \methodname{} addresses the decoupled tasks through multiple specialized projections, each designed to model specific aspects of visual information processing. For \textit{Key Object Segmentation}, we train a text-driven video decoder to generate masks for the key object in a video, guided by the key object class name. For \textit{Concept Recognition}, we train the brain model to recognize semantic concepts in the video through a multi-label classification task. For \textit{Scene Description}, we teach the brain model to describe the video by decoding text embeddings into captions. Finally, for \textit{Blurry Video Reconstruction}, the brain model is guided to align with the latent space of the Stable Diffusion's VAE~\cite{rombach2022ldm}, using the same video decoder as in segmentation but replacing the segmentation head with a reconstruction head.
During inference, the decoupled task outputs of \methodname{} serve as robust conditioning signals for a pre-trained text-to-video (T2V) diffusion model, enabling high-fidelity video reconstruction with improved temporal smoothness and semantic consistency.

Extensive experiments show that \methodname{} outperforms state-of-the-art baselines, especially on video-based metrics. It improves spatiotemporal consistency by 0.196 (26.6\%) and semantic metrics by 0.042 (19.1\%) on average. The proposed sub-tasks enhance shape, location, and semantic extraction, and their decoupling boosts brain activity decoding. Mapping sub-task projections onto the brain's visual regions reveals functional alignment with the human visual cortex.

%% file: sec/2_related_works.tex
\section{Related Works}

\subsection{fMRI-to-Video Reconstruction}
Recently, the decoding of original visual stimuli (e.g., images and videos) from brain functional magnetic resonance imaging (fMRI) has garnered significant attention. This interest stems from its crucial role in uncovering the intricate mechanisms of the human brain~\cite{takagi2023high}, as well as its remarkable potential to facilitate novel clinical assessment methods and advance brain-computer interface applications~\cite{wen2018neural}. Early exploration on this topic mainly focuses on how to decode fMRI signals into concepts~\cite{horikawa2017firstfmri}, and in recent years, the reconstruction of static images~\cite{takagi2023high,mai2023unibrain,gong2023lite-mind,bao2024willsaligner} and dynamic videos~\cite{gong2025neuroclips,li2024enhancing} have also witnessed some progresses. The typical paradigm consists of two crucial steps: 1) Align the representations of fMRI and visual stimuli and map them both to the CLIP~\cite{radford2021clip} embedding space; 2) utilize generative models (\textit{e.g.}, diffusion model~\cite{rombach2022high}, generative adversarial network~\cite{goodfellow2020generative}) to reconstruct the visual stimuli. More recently, some researchers started to explore the more challenging reconstruction task from fMRI to dynamic video stimuli. For example, MinD-Video~\cite{chen2024cinematic} successfully reconstructs videos from fMRI at a relatively low FPS. NeuroClips~\cite{gong2025neuroclips} attempted to improve the semantic accuracy and video smoothness with a stronger semantic alignment paradigm and a more detailed perception capturing.
The main reconstruction process of these methods relies on aligning the brain model with CLIP features \textit{implicitly}, \textit{i.e.}, aligning in the hidden feature space, which is not robust. The reason is that the CLIP hidden feature space is highly semantic, but the fMRI voxels are encoded with multi-granularity information.

\subsection{Diffusion Models for Video Synthesis}
Recently, the diffusion model has garnered significant attention in the generative artificial intelligence communities. The breakthrough came with Ho et al.~\cite{ho2020ddpm}, who introduced the Denoising Diffusion Probabilistic Model (DDPM), demonstrating high-quality image synthesis comparable to GANs. In the static image generation realm, after DDPM, DALLE-2~\cite{ramesh2022hierarchical} incorporate CLIP to improve the performance of text-to-image synthesis. Stable Diffusion~\cite{rombach2022ldm} improved generation efficiency by performing the diffusion process in the latent space of VQVAE~\cite{van2017vqvae}. In the more difficult realm, \textit{i.e.}, dynamic video synthesis, the typical pipeline is to incorporate an additional temporal module to improve the smoothness of the generated videos. For example, Blattmann et al.~\cite{blattmann2023alignyourlatent} introduced a Temporal Autoencoder Finetuning paradigm, which achieved high-resolution and high-fps video generation. More recently, AnimateDiff~\cite{guo2023animatediff} introduced an innovative temporal motion module that seamlessly extends existing diffusion models for static image generation into dynamic video generation.

%% file: sec/3_method.tex
\section{Methodology}

\begin{figure*}[tb]
    \centering
    \includegraphics[width=1\linewidth]{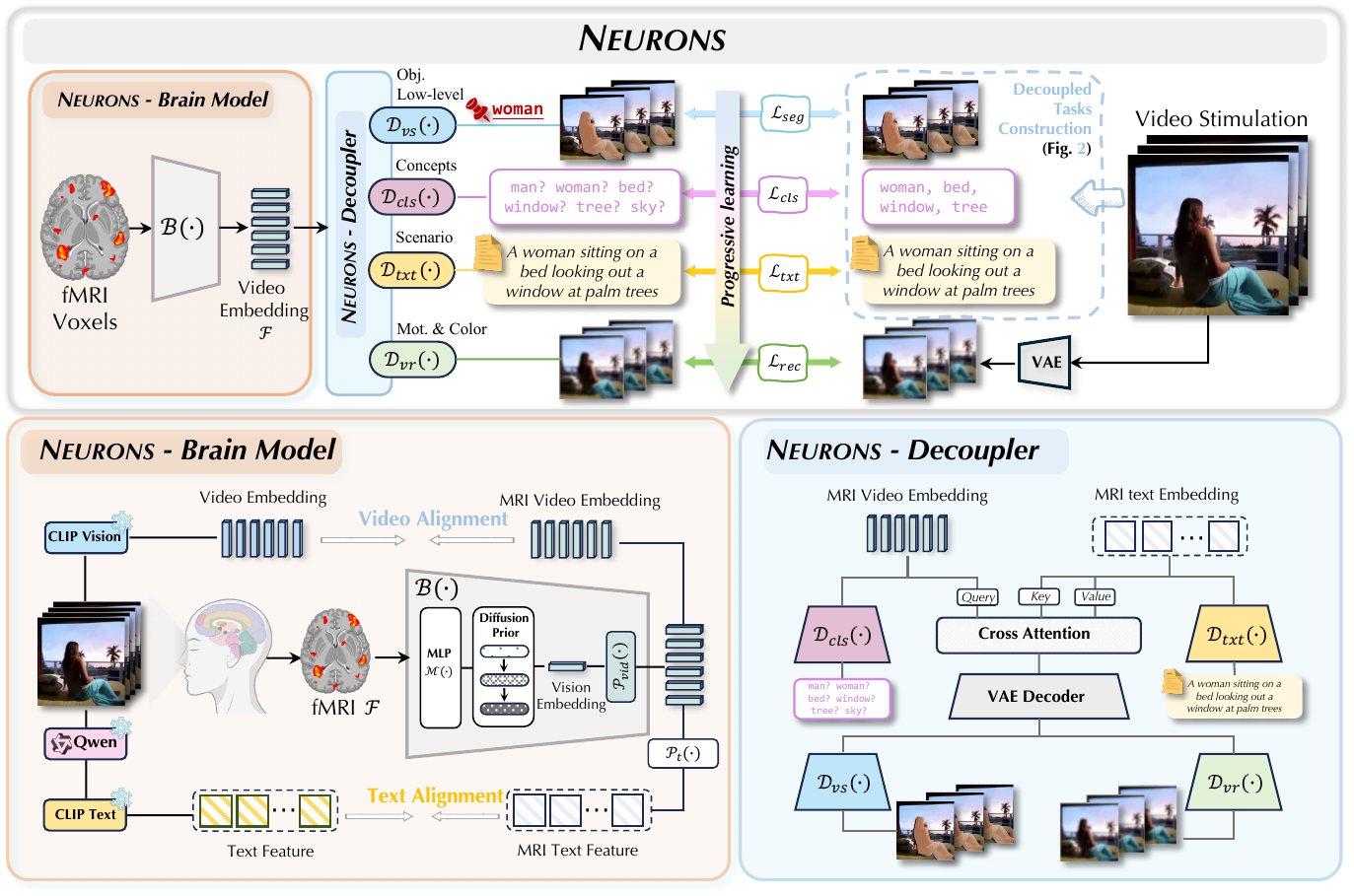}
    \caption{The overall framework of \methodname{} comprises two primary components: the Brain Model (\S\ref{sec:brain_model}) and the Decoupler (\S\ref{sec:decoupler}). The Brain Model is responsible for mapping fMRI representations to motion embeddings, while the Decoupler decouples these embeddings into distinct explicit sub-tasks, including key object segmentation, concept recognition, scene description, and blurry video reconstruction. These four tasks are progressively learned by dynamically adjusting the loss weights. The outputs generated from these sub-tasks are subsequently utilized to reconstruct the final videos during the inference stage, as illustrated in Fig.~\ref{fig:infer}.}
    \label{fig:framework}
\end{figure*}

The overall framework of \methodname{} is illustrated in Fig.~\ref{fig:framework}. \methodname{} simulates the functional behaviors of neurons in the human visual cortex by learning four decoupled tasks, the construction of which is detailed in \S \ref{sec:DTC}. The technical core of \methodname{} consists of three main components: (1) a \textit{Brain Model} that maps fMRI representations to motion embeddings (\S \ref{sec:brain_model}), (2) a \textit{Decoupler} that disentangles the training of motion embeddings into progressive and explicit sub-tasks (\S \ref{sec:decoupler}), and (3) an \textit{aggregated video reconstruction pipeline} that integrates the outputs of all sub-tasks to guide high-quality video reconstruction (\S \ref{sec:infer}).

\begin{figure}[tb]
    \centering
    \includegraphics[width=1\linewidth]{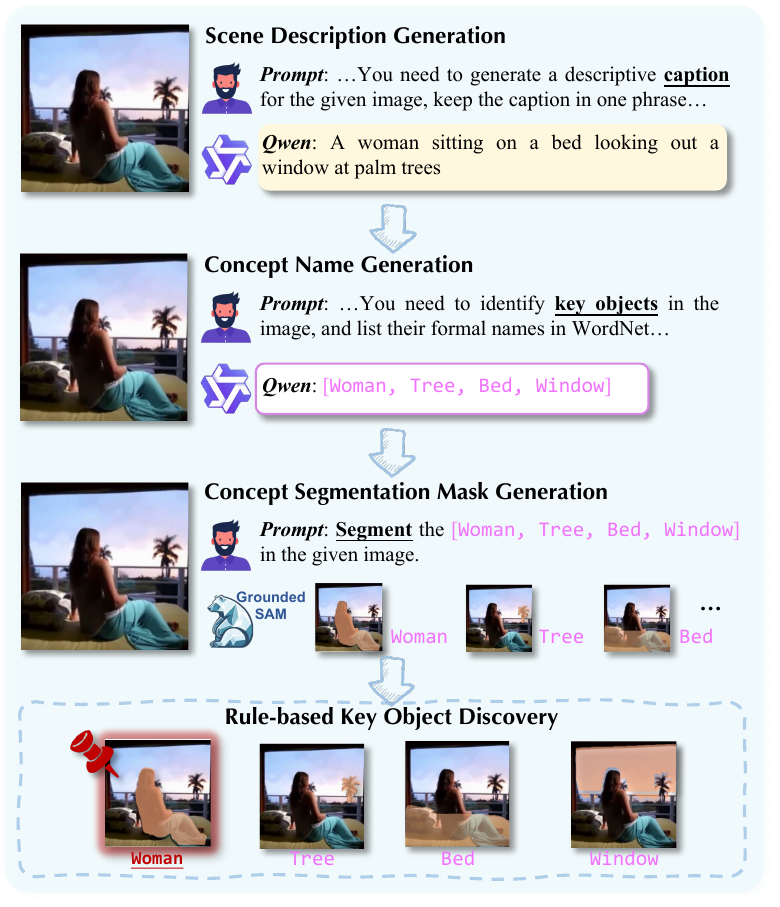}
    \caption{Decoupled Tasks Construction.}
    \label{fig:setup}
\end{figure}

\subsection{Decoupled Tasks Construction} \label{sec:DTC}
We respectively carry out four task-setting build-up processes: scene description generation, concept name generation, concept segmentation mask generation, and rule-based key object discovery, as illustrated in Fig.~\ref{fig:setup}.

\noindent\textbf{Scene Description Generation.}
To build the task setting for the scene description task, we need to generate a descriptive caption for each frame. The key tool we utilize is the Qwen2.5-VL-72B-Instruct~\cite{qwen}. Firstly, we input the frame image into Qwen, and instruct it to provide a detailed one-phrase caption for the image, such as ``\texttt{a woman sitting on a bed looking out a window at palm trees}". The detailed instruction prompts for scene description generation are illustrated in Supplementary \S~\ref{sec:supp_prompts}.

\noindent\textbf{Concept Name Generation.}
To build the task setting for the concept recognition, we need to generate the concept names of the key objects in each frame. Similarly, we first input the frame into Qwen. Then, we instruct the Qwen to recognize the primary objects (\textit{e.g.}, [\textquotesingle\texttt{woman}\textquotesingle, \textquotesingle\texttt{bed}\textquotesingle, \textquotesingle \texttt{tree}\textquotesingle, ...]) in Fig.~\ref{fig:setup}) in the frame. The primary objects will then be categorized into 51 concepts ([\textquotesingle\texttt{animal}\textquotesingle, \textquotesingle\texttt{human}\textquotesingle, \textquotesingle\texttt{plant}\textquotesingle, \textquotesingle\texttt{furniture}\textquotesingle, ...]), which are summarized from the WordNet~\cite{wordnet}. The output format is a concept name list (\textit{e.g.}, [\textquotesingle\texttt{human}\textquotesingle, \textquotesingle\texttt{furniture}\textquotesingle, \textquotesingle\texttt{plant}\textquotesingle] for Fig.~\ref{fig:setup}) that covers all primary objects in the frame. The detailed instruction prompts for concept name generation and the 51-concept list are in Supplementary \S~\ref{sec:supp_prompts}.

\noindent\textbf{Concept Segmentation Mask Generation.}
Now, we utilize the object list for each frame to generate segmentation masks for the primary objects in each frame. We input each frame along with its object list into Grounded-SAM~\cite{ren2024groundedSAM} and instruct it to generate a binary mask for each object using the object names from the list as textual prompts.

\noindent\textbf{Rule-based Key Object Discovery.}
We propose a multi-criteria approach to identify key objects in video sequences by integrating motion dynamics, object size, and semantic importance. Background categories (\textit{e.g.}, sky, ocean) are excluded using predefined labels. For remaining objects, inter-frame displacement is calculated, with higher weights for priority categories like humans and animals (\textit{e.g.}, person, dog, cat) to reflect semantic importance. Objects are ranked by weighted displacement, and excessively large ones (\textit{i.e.}, $>$50\% of the image area) are filtered out. Priority categories are selected first; otherwise, top-ranked objects by motion dynamics are chosen. If all objects are background, the largest one is selected as a fallback to ensure at least one key object is identified.

\subsection{{\large\methodname{}} - Brain Model}\label{sec:brain_model}

The brain model aims to encode fMRI voxels into video embeddings. Its training serves as a pre-training stage to produce suitable video embeddings, preparing for subsequent decoupled tasks.
After preprocessing, a video is split into several clips at two-second intervals (\textit{i.e.}, fMRI time resolution), each video clip $y_c$ has six frames $y_c\in\mathbb{R}^{B\times F\times C\times H\times W}$ where $F$ the number of frames and $F=6$.
$x_c$ is the corresponding fMRI signal of video clip $c$. 
We use the same pre-trained MindEyeV2~\cite{scotti2024mindeye2} (details are illustrated in Supplementary \S \ref{sec:supp_BMdetail}).
Previous approaches like NeuroClips directly align the image embeddings from the backbone, overlooking temporal relationships between consecutive frames. Instead, we use a motion projection $\mathcal{P}_{vid}(\cdot)$ to map the image embeddings $e^i\in\mathbb{R}^{B\times N \times C}$ to spatial-temporal space $e^v\in\mathbb{R}^{B\times F\times N \times C}$.
We use the same bidirectional contrastive loss as in~\cite{scotti2024mindeye2,gong2025neuroclips} called BiMixCo, denoted as $\mathcal{L}_\mathrm{CLIP_v}$, which combines MixCo and contrastive loss, to align $e^v$ with the target video embeddings of CLIP vision encoder, detailed in Supplementary \S \ref{sec:supp_BMdetail}.
Then, we embed $e^v$ to
obtain the text embedding $e^t$. A contrastive learning loss between CLIP text embedding and $e^t$ is adopted to train the additional text modality. The contrastive loss serves as the training loss $\mathcal{L}_\mathrm{CLIP_t}$ of this process, similar to $\mathcal{L}_\mathrm{CLIP_v}$, omitted here.
This pre-alignment provides feasible vision and text features which are then fed into the Decoupler for decoupled task learning.

\begin{figure}
    \centering
    \includegraphics[width=1\linewidth]{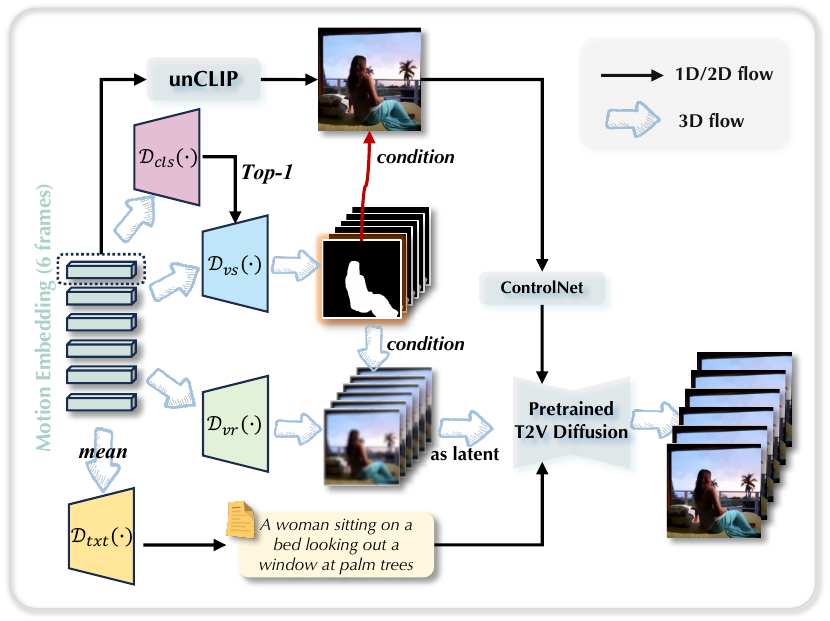}
    \caption{Inference pipeline of \methodname{}. We use the generated key object masks as condition to control image generation and blurry video generation, and the generated scene description as text prompt to T2V Diffusion.}
    \label{fig:infer}
\end{figure}

\subsection{{\large\methodname{}} - Decoupler: Decoupled Explicit Tasks Learning}
\label{sec:decoupler}
Starting from a key object segmentation task that requires minimum semantic information, we \textit{progressively} teach \methodname{} to learn more complex semantics through \textit{concept recognition task} (to recognize semantic concepts), \textit{scene description task} (to describe the video by decoding text embeddings into captions), and \textit{blurry video reconstruction task} (to learn motion information and color distribution).
During inference, the decoupled task outputs of \methodname{} serve as robust conditioning signals for a pre-trained text-to-video (T2V) diffusion model, enabling high-fidelity video reconstruction with improved temporal smoothness and semantic consistency.

\noindent\textbf{Key Object Segmentation.} The capacity of humans to retain detailed memories of videos is inherently limited. Research has shown that individuals tend to focus on \textit{key objects} and salient features within a scene, often overlooking finer details~\cite{keyobj_judd2009learning}. This selective attention mechanism provides a foundational strategy to reduce the complexity of the learning process. By initially focusing on key objects, we can progressively introduce more complex semantic concepts and motion information.
To complete this task, we design a text-driven video decoder that is based on the VAE video decoder~\cite{von-platen-etal-2022-diffusers}. The decoder takes video embeddings $e^v$ and text embeddings $e^t$ as inputs. For this task, the text embeddings are obtained by encoding the category name of the key object with the CLIP text encoder. Then, a cross-attention module is used to activate certain patches of $e^v$ (as query $Q$) corresponding to $e^t$ (as key $K$ and value $V$): ${e}^{seg}  = \mathrm{softmax}(\frac {Q K^\top} {\sqrt{d}}) \cdot V$,
the activated $e^{seg}$ is upsampled to a higher resolution for pixel-level recognition. Finally, we use a simple segmentation head $\mathcal{D}_{vs}(\cdot)$ to generate binary segmentation masks $y_{seg}$ for the key object in the video. The objective function of this task is a binary cross-entropy loss $\mathcal{L}_{seg}$, detailed in Supplementary \S \ref{sec:supp_seg}.

\noindent\textbf{Concept Recognition.}
Despite the detailed visual information we learn from the above segmentation and scene description tasks, it is equally important to understand and recognize the semantic concept in the frame image to prevent semantic shifts. To this end, we incorporate a concept recognition task by additionally involving a multi-label classifier $\mathcal{D}_{cls}(\cdot)$ to recognize the key concepts inside the frame image using the vision embedding of MRI. Specifically, we impose the cross-entropy loss between the prediction and the GT concept list (Sec.~\ref{sec:DTC}):
\begin{equation}
    \label{eq:cls}
    \mathcal{L}_{cls} = \mathcal{L}_{ce}(\mathcal{D}_{cls}(\bar{e^v}), \mathcal{C}),
\end{equation}
where $\bar{e^v}$ represents the mean of ${e^v}$ computed along the frame axis and $\mathcal{C}$ denotes the GT concept list.

\noindent\textbf{Scene Description.}
Next, with the ability to identify key objects and complementary concepts, the brain model is taught to comprehensively describe the overall visual scene. To achieve that, we incorporate a scene description task, which aims to generate a descriptive caption for each frame of the video. Specifically, we incorporate and finetune a pre-trained text decoder $\mathcal{D}_{txt}(\cdot)$, which takes the text embeddings of the fMRI as input to generate the caption text. Herein, we utilize the GPT-2~\cite{gpt2} as our text decoder. We train the text decoder through prefix language modeling. Specifically, given a ground-truth (GT) caption token sequence $\mathcal{S}={s_0,s_1,...,s_{|\mathcal{S}|}}$ and the corresponding text embedding of the MRI $e^t$, our text decoder $\mathcal{D}_{txt}(\cdot)$ learns to reconstruct $\mathcal{S}$ regarding $e^t$ as the prefix. The objective function $\mathcal{L}_{txt}$ can be described as follows:
\begin{equation}
    \label{eq:txt}
    \mathcal{L}_{txt} = -\frac{1}{|\mathcal{S}|} \sum^{|\mathcal{S}|}_{i=1} \log \mathcal{D}_{txt}(s_i|s_{<i}, e^t),
\end{equation}
where $s_i$ denotes the $i^{th}$ token in GT token sequence $\mathcal{S}$.

\noindent\textbf{Blurry Video Reconstruction.}
Finally, the blurry video reconstruction task enables the model to learn the color distribution and capture motion information.
We adopt the same VAE decoder used in the key object segmentation task while replacing the segmentation head $\mathcal{D}_{vs}(\cdot)$ with a reconstruction head $\mathcal{D}_{vr}(\cdot)$
Take the video embeddings $e^t$ as inputs, the text-driven video decoder with $\mathcal{D}_{vr}(\cdot)$ generates blurry video $y^{rec}_c\in\mathbb{R}^{B\times F \times C \times \frac {H} {8} \times \frac {W} {8}}$.
Then we map ${y_c}$ to the latent space of Stable Diffusion's VAE to obtain the latent embeddings ${y'_c}$. We adopt mean absolute error (MAE) loss to train this sub-task, the overall loss can be described as:
\begin{equation}
\label{contrastive}
\mathcal{L}_{rec} = \frac{1}{F} \sum^{F}_{i=1} | y^{rec}_{c,i} - y'_{c,i} |.
\end{equation}

\noindent\textbf{Progressive Learning Strategy and Overall Training Loss.}
To dynamically adjust the weights of different loss functions during training, we propose a logarithmic weight scheduling strategy. This strategy ensures that the weight of each loss function follows a smooth curve, starting from 1, increasing to 10, and then decreasing back to 1 over a predefined period. The weight scheduling is applied to four loss functions, each with a staggered period to ensure balanced training.
For epoch $E$, batch $B$, and batches per epoch $N_B$, the total batches in a period $P$ is $T = P \cdot N_B$.
The current batch's position within the period starting at epoch $S$ is $C = (E - S) \cdot N_B + B$.
The weight $w$ is computed as $w = 1 + 9 \cdot \left| \sin\left(\frac{C}{T} \cdot \pi\right) \right|$.
If $E$ is outside the period $[S, S + P)$, $w = 1$.
At each batch step, the weights for all four loss functions are computed using the above formula, with their periods staggered, which ensures smooth transitions, as shown in Supplementary Fig.~\ref{fig:weights}.
The overall training loss can be defined as:
\begin{equation}
    \label{semantic-reconstruction-loss}
    \mathcal{L}_{total}=w_1\mathcal{L}_{seg}+w_2\mathcal{L}_{cls}+w_3\mathcal{L}_{txt} +w_4\mathcal{L}_{rec}.
\end{equation}

\subsection{Inference: Aggregated Video Reconstruction}\label{sec:infer}
Following NeuroClips~\cite{gong2025neuroclips}, we use a pre-trained T2V diffusion model~\cite{guo2023animatediff} for inference, guided by a control image, blurry video, and text description. As shown in Fig.~\ref{fig:infer}, we prepare these inputs using outputs from decoupled tasks: the control image is generated from a frame via unCLIP~\cite{ramesh2022hierarchical}, while concepts and text descriptions are derived from $\mathcal{D}_{cls}$ and $\mathcal{D}_{txt}$. The top-1 concept and text embeddings guide video mask generation via $\mathcal{D}_{vs}$ and blurry video creation via $\mathcal{D}_{vs}$, respectively. To emphasize the key object, we rescale its binary masks to $[0.5, 1]$ and multiply them to condition both the control image and blurry video, ensuring its prominence.

%% file: sec/4_exp.tex
\section{Experiments}

\subsection{Experimental Setup}
In this study, we performed fMRI-to-video reconstruction experiments using the open-source fMRI-video dataset (cc2017 dataset\footnote{\url{https://purr.purdue.edu/publications/2809/1}})~\cite{wen2018neural}. For each subject, the training and test video clips were presented 2 and 10 times, respectively, and the test set was averaged across trials. The dataset consists of a training set containing 18 8-minute video clips and a test set containing 5 8-minute video clips. The MRI (T1 and T2-weighted) and fMRI data (with 2s temporal resolution) were collected using a 3-T MRI system. Thus there are 8640 training samples and 1200 testing samples of fMRI-video pairs.  The data pre-processing follows NeuroClips~\cite{gong2025neuroclips}, which is detailed in Supplementary \S\ref{sec:supp_preprocessing}.
Implementation details are provided in Supplementary \S\ref{sec:supp_inplement}.

\noindent\textbf{Evaluation Metrics.}
We conduct the quantitative assessments primarily focusing on video-based metrics. 
We evaluate videos at the semantic level and spatiotemporal(ST)-level. For semantic-level metrics, a similar classification test (total 400 video classes from the Kinetics-400 dataset~\cite{kay2017kinetics}) is used above, with a video classifier based on VideoMAE~\cite{tong2022videomae}. 
For spatiotemporal-level metrics that measure video consistency, we compute CLIP image embeddings on each frame of the predicted videos and report the average cosine similarity between all pairs of adjacent video frames, which is the common metric CLIP-pcc in video editing~\cite{clippcc}.
Additionally, we assess frames at both pixel and semantic levels. For pixel-level evaluation, we use SSIM and PSNR. For semantic-level evaluation, we employ an N-way top-K accuracy classification test (1,000 ImageNet classes). The test compares ground truth (GT) classification results with predicted frame (PF) results using an ImageNet classifier. A trial is successful if the GT class is among the top-K probabilities (we used top-1) in the PF results, selected from N random classes, including the GT. The success rate is based on 100 repeated tests.

\begin{table}[!t]
\centering
  \caption{Quantitative comparison of \methodname{} reconstruction performance against other methods (Video-based). Bold font signifies the best performance, while underlined text indicates the second-best performance. MinD-Video and \textit{NeuroClips} are both results averaged across all three subjects, and the other methods are results from subject 1. Results of baselines are quoted from~\cite{lu2024animate,gong2025neuroclips}.}
\resizebox{\linewidth}{!}{
  \begin{tabular}{lccc}
    \toprule
    \multirow{2}{*}{Method} & \multicolumn{2}{c}{Semantic-level} & ST-level \\
    \cmidrule(lr){2-3} \cmidrule(lr){4-4}
    & 2-way$\uparrow$ & 50-way$\uparrow$ & CLIP-pcc$\uparrow$ \\
    \midrule

    Wen~\cite{wen2018neural} & - & 0.166\scriptsize{$\pm0.02$} & - \\
    Wang~\cite{wang2022reconstructing} & 0.773\scriptsize{$\pm0.03$} & - & 0.402\scriptsize{$\pm0.41$} \\
    Kupershmidt~\cite{kupershmidt2022penny} & 0.771\scriptsize{$\pm0.03$} & - & 0.386\scriptsize{$\pm0.47$} \\
    MinD-Video~\cite{chen2024cinematic} 
    & \underline{0.839\scriptsize{$\pm0.03$}} & {0.197\scriptsize{$\pm0.02$}} & {0.408\scriptsize{$\pm0.46$}} \\

    MindAnimator~\cite{lu2024animate} 
    & {0.830\scriptsize{$\pm0.03$}} &- & {0.428\scriptsize{$\pm0.03$}} \\
    NeuroClips~\cite{gong2025neuroclips} 
    & {0.834\scriptsize{$\pm0.03$}} & \underline{0.220\scriptsize{$\pm0.01$}} & \underline{0.738\scriptsize{$\pm0.17$}} \\
    
    \rowcolor{cyan!10} 
    \textbf{\methodname{}} 
    & \textbf{0.863\scriptsize{$\pm0.03$}} & \textbf{0.262\scriptsize{$\pm0.01$}} & \textbf{0.934\scriptsize{$\pm0.17$}} \\
    \midrule
    \rowcolor{cyan!10} 
    subject 1 & {0.862}\scriptsize{$\pm0.02$} & {0.254}\scriptsize{$\pm0.02$} & {0.932}\scriptsize{$\pm0.04$} \\
    \rowcolor{cyan!10} 
    subject 2 & {0.860}\scriptsize{$\pm0.03$} & {0.252}\scriptsize{$\pm0.02$} & {0.933}\scriptsize{$\pm0.04$} \\
    \rowcolor{cyan!10} 
    subject 3 & {0.868}\scriptsize{$\pm0.02$} & {0.278}\scriptsize{$\pm0.02$} & {0.937}\scriptsize{$\pm0.02$} \\
    \bottomrule
  \end{tabular}}
\label{tab:video}
\end{table}

\begin{figure}[htb]
    \centering
    \includegraphics[width=1\linewidth]{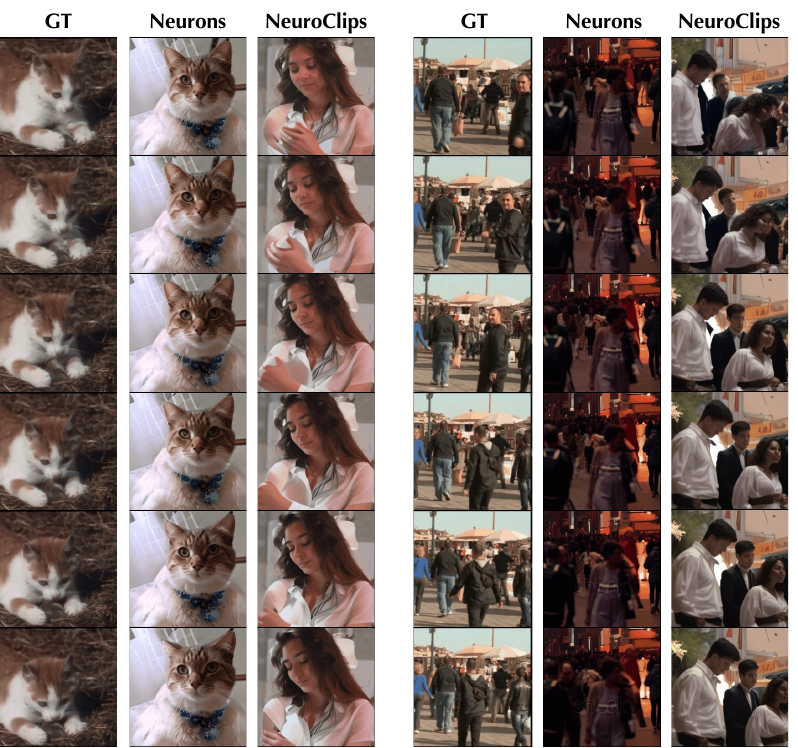}
    \caption{Qualitative comparison between \methodname{} and previous SOTA method, NeuroClips~\cite{gong2025neuroclips}.}
    \label{fig:SOTA_vis}
\end{figure}

\subsection{Comparison with State-of-the-Art Methods}
In this section, we compare \methodname{} with previous video reconstruction methods on the cc2017 dataset both quantitatively and qualitatively. 
In quantitative metrics, our proposed method demonstrates superior performance across all video-based metrics compared to existing methods, as shown in Table 1. Semantic-level 2-way accuracy achieves 0.863, surpassing the previous best result (0.839 by MinD-Video) by a relative margin of 2.9\%. Notably, our method significantly outperforms state-of-the-art methods in ST-level CLIP-pcc, attaining 0.934, which represents a 26.6\% improvement over NeuroClips (0.738). Furthermore, our method achieves the highest 0.262 in the 50-way accuracy, outperforming NeuroClips (0.220) by 19.1\%. The consistent superiority across all metrics, coupled with lower standard deviations, underscores the robustness of our approach. These results validate that our method effectively bridges semantic and spatiotemporal feature learning, setting a new benchmark for video understanding tasks.
Moreover, we also provided a frame-based comparison, which is detailed in Supplementary \S \ref{sec:supp_frame_comp} and Table~\ref{tab:frame}.
For qualitative comparison, as can be seen in Fig.~\ref{fig:SOTA_vis}, compared to NeuroClips, our \methodname{} generates videos with higher quality, more precise semantics (\textit{e.g.}, people, turtles, and airplanes), and smoother movements. 
Furthermore, thanks to the decoupled tasks, \methodname{} has a better capability of reconstructing more complex scenes, such as a crowd of people. More visualizations can be found in Supplementary \S\ref{sec:supp_morevis}.

\subsection{Ablation Studies}
In this section, we evaluate the key components of \methodname{}, including the four decoupled sub-tasks, the progressive learning strategy, and the aggregated video reconstruction. 
The quantitative results are in Table \ref{tab:ablation}, where all the results of the ablation experiments are from subject 1. 
Due to the hierarchical function of the proposed methods, we gradually add components starting from the brain model with $\mathcal{L}_{rec}$, which is the minimum input requirement of the NeuroClips inference pipeline.
The results indicate that the inclusion of $\mathcal{L}_{seg}$ and $\mathcal{L}_{cls}$ significantly improves the model's performance, as evidenced by the increase in 2-way and 50-way metrics. The addition of $\mathcal{L}_{txt}$ further enhances the semantic-level understanding, leading to a notable improvement in the CLIP-pcc score. The progressive learning strategy and aggregated video reconstruction, when combined with all other components, yield the highest performance across all metrics, achieving the best 2-way and 50-way scores of 0.862 and 0.254, respectively. Qualitative analysis is detailed in Supplementary \S \ref{sec:supp_ablation}.

\begin{table}[t]
  \centering
  \caption{Ablations on the key components of \methodname{} on video-based metrics, and all results are from subject 1. `PL' denotes the progressive learning strategy and `AVR' stands for aggregated video reconstruction.}
  \resizebox{\linewidth}{!}{
  \begin{tabular}{cccccccccc}
    \toprule
    Brain & \multicolumn{6}{c}{Key Components}  & \multicolumn{2}{c}{Semantic-level} & ST-level \\
    \cmidrule(lr){2-7} \cmidrule(lr){8-9} \cmidrule(lr){10-10} Model &
    $\mathcal{L}_{seg}$ & $\mathcal{L}_{cls}$ & $\mathcal{L}_{txt}$ & $\mathcal{L}_{rec}$ & PL & AVR & 2-way$\uparrow$ & 50-way$\uparrow$ & CLIP-pcc$\uparrow$ \\
            
    \midrule

    \ding{51} & &  & &\ding{51} & & & 0.814 & 0.164 & 0.894 \\
    
    \ding{51} &\ding{51} & & &\ding{51} & & & 0.834 & 0.225 & 0.926 \\
     
    \ding{51} &\ding{51} &\ding{51} & &\ding{51} &  & & {0.836} & {0.234} & 0.911 \\

    \ding{51} &\ding{51} &\ding{51} &\ding{51} &\ding{51} & & &{0.847} & {0.213} & 0.923 \\

    \ding{51} &\ding{51} &\ding{51} &\ding{51} &\ding{51} &\ding{51} & & 0.856 & 0.235 & \textbf{0.937} \\
     
    \midrule
    \rowcolor{cyan!10}
    \ding{51} &\ding{51} &\ding{51} &\ding{51} &\ding{51} &\ding{51} &\ding{51} & \textbf{0.862} & \textbf{0.254} & {0.932} \\
    
     \bottomrule
    \end{tabular}}
  \label{tab:ablation}
\end{table}

\section{Explicit Evaluation of Decoupled Tasks}
Each decoupled task provides certain improvements to the final reconstruction results. In this section, we explicitly evaluate the outputs of decoupled tasks, \textit{i.e.}, the accuracy of segmentation, classification, and description generation.

\noindent\textbf{Key Object Segmentation.}
We conducted a comprehensive evaluation of the predicted key object masks. Although the overall Dice score is relatively low (35.63\%), the localization of key objects is often accurate in many cases. The lower Dice score can be attributed to the presence of numerous out-of-distribution categories in the test set, as highlighted in Table~\ref{tab:all_class_accuracy}.
In Fig.~\ref{fig:seg_vis}, we present several successful examples where the model achieves precise localization of the key object. These results demonstrate the model's ability to effectively refine the shape and position of objects in blurry video frames.
\begin{figure*}
    \centering
    \includegraphics[width=\linewidth]{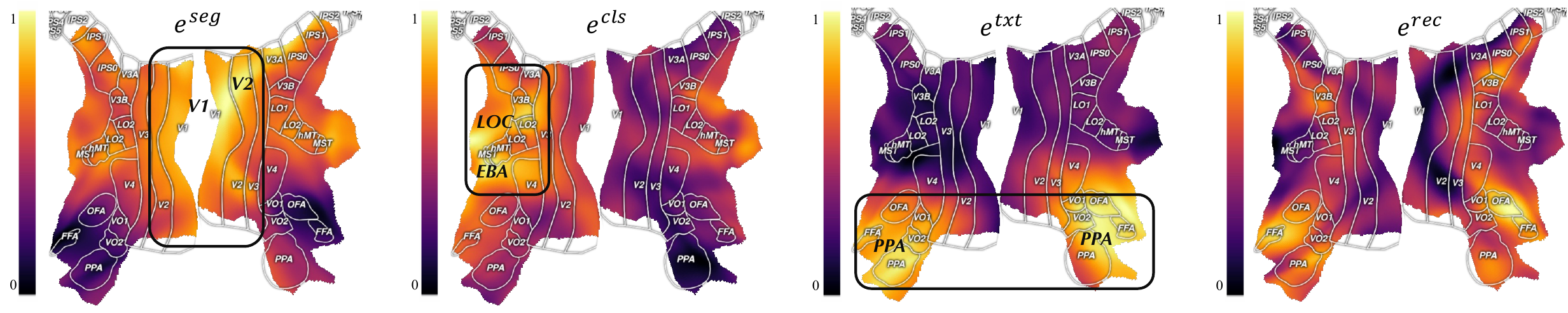}
    \caption{Mapping decoupled embeddings to the brain map~\cite{yang2024braindecodesdeepnets} shows consistency with the design ethos in Fig.~\ref{fig:abstract}: object segmentation captures low-level features (V1, V2), concept recognition identifies body parts (EBA), and objects (LOC), scene description processes scenes (PPA) and higher-level representations, and blurry video reconstruction integrates global but less low-level information.}
    \label{fig:interpretation}
\end{figure*}

\begin{figure}[t]
        \centering
        \animategraphics[width=\linewidth]{8}{videos/video_8/}{01}{06}
        \animategraphics[width=\linewidth]{8}{videos/video_150/}{01}{06}
        \animategraphics[width=\linewidth]{8}{videos/video_425/}{01}{06}
    \caption{From left to right, the videos are as follows: the GT (ground truth) video, the blurry video before and after applying the segmentation mask condition, the segmentation GT, and the predicted segmentation masks. \textit{Best viewed with Acrobat Reader. Click the images to play the animation clips.}}
    \label{fig:seg_vis}
\end{figure}

\noindent\textbf{Concept Recognition.}
We evaluated multi-class classification accuracy to assess the concept recognition task, ensuring precise and meaningful semantic representations from the visual cortex. As shown in Table~\ref{tab:combined_results}, \methodname{} achieves satisfactory accuracy on common categories (0.735 on human), which indicates the ability of \methodname{} to preserve semantic integrity and prevent conceptual deviations.

\begin{table}[t]
\centering
\caption{(Top) Evaluation results for concept recognition. Only the top-5 classes are displayed here, Fullr results in Supplementary \S \ref{sec:supp_cls}. (Bottom) Evaluation of scene description at both the sentence level and verb level.}
\resizebox{0.85\linewidth}{!}{
\begin{tabular}{c|ccccc}
\toprule
\multicolumn{6}{c}{\textbf{ Concept Recognition Evaluation}}\\ \midrule
Rank & 1 & 2 & 3 & 4 & 5 \\ 
Class Name & human & food & animal & water body & fish \\ \midrule
\rowcolor{cyan!10} 
Accuracy & 0.735 & 0.600 & 0.450 & 0.310 & 0.292 \\ \bottomrule
\end{tabular}}

\resizebox{\linewidth}{!}{
\begin{tabular}{c|ccccc|c}
\toprule
\multicolumn{7}{c}{\textbf{Scene Description Evaluation}}\\ \midrule
\multirow{2}{*}{Metric} & \multicolumn{5}{c|}{Sentence-level} & \multirow{2}{*}{Verb Acc}\\
& {Bleu\_1} & {Bleu\_2} & {Bleu\_3} & {Bleu\_4} & {CIDEr} &  \\  \midrule

NeuroClips & 0.227 & 0.096 & 0.042 & 0.022 & 0.156 & 0.1158\\
\rowcolor{cyan!10} 
\methodname{} & \textbf{0.238} & \textbf{0.105} & \textbf{0.056} & \textbf{0.036} & \textbf{0.239} & \textbf{0.2425}\\ 
\bottomrule
\end{tabular}}
\label{tab:combined_results}
\end{table}

\noindent\textbf{Scene Description.}
NeuroClips~\cite{gong2025neuroclips} generates text prompts for the T2V diffusion model by captioning the generated keyframes using BLIP. However, this approach lacks robustness due to inaccuracies in the generated keyframes, which often hinder the precise identification of semantic content, as demonstrated in Table~\ref{tab:combined_results}. We conduct a comprehensive comparison between the descriptions generated by NeuroClips and our method. The results reveal that \methodname{} produces more accurate scene descriptions at the sentence level, particularly in capturing verbs that reflect better motion information. Further details in Supplementary \S \ref{sec:supp_capverb_detail}.

\section{Interpretation Results}
\label{interpretation}
Inspired by the human visual cortex, we designed our \methodname{} with four different decoupled tasks. In order to validate our initial inspiration, we map the projected embeddings of each decoupled task to a brain map with~\cite{yang2024braindecodesdeepnets}, a visualization tool that uncovers functional correlations between \methodname{}'s tasks and distinct regions of the visual cortex. As shown in Figure~\ref{fig:interpretation}, the embeddings align with specific regions: $e^{seg}$ with V1, V2, and MT (due to video-based segmentation), $e^{cls}$ with V4, EBA and LOC, $e^{txt}$ with PPA and FFA. Notably, $e^{rec}$ highlights V3 areas while also encoding richer semantic information. This alignment with various visual cortex regions validates our design principles and underscores the biological plausibility of our approach. More details are in Supplementary \S\ref{sec:Interpretation}.

%% file: sec/5_conclusion.tex
\section{Conclusion}
We present \methodname{}, a novel framework inspired by the hierarchical organization of the human visual cortex, designed to enhance the fidelity and interpretability of fMRI-to-video reconstruction. By decomposing the learning process into four specialized tasks, we effectively emulate the functional specialization of the human visual system. This hierarchical approach enables the framework to capture a broad spectrum of video content, ranging from low-level visual features to high-level semantic concepts.
Our comprehensive experimental results demonstrate that \methodname{} surpasses previous methods across both video-based and frame-based evaluation metrics. The framework achieves marked improvements in spatiotemporal consistency, semantic accuracy, and pixel-level reconstruction quality.
The decoupled task structure not only enhances reconstruction performance but also provides interpretable insights into the functional alignment between the model and the human visual cortex. By mapping the outputs of each task to corresponding brain regions, we observed a strong correlation with the hierarchical processing of visual information in the human brain, which further highlights its potential for brain-computer interfaces and clinical applications.

\section*{Acknowledgements}
This work was supported by a research grant from the Joint Research Scheme (JRS) under the National Natural Science Foundation of China (NSFC) and the Research Grants Council (RGC) of Hong Kong (Project No. N\_HKUST654/24), as well as a grant from the Research Grants Council of the Hong Kong Special Administrative Region, China (Project No. R6005-24).

%% file: sec/X_suppl.tex
\clearpage
\setcounter{page}{1}
\setcounter{section}{0}
\setcounter{figure}{0}
\setcounter{table}{0}
\renewcommand\thesection{\Alph{section}}
\renewcommand\thetable{\alph{table}}
\renewcommand\thefigure{\alph{figure}}
\maketitlesupplementary

\begin{figure*}[ht]
    \centering
    \includegraphics[width=1\linewidth]{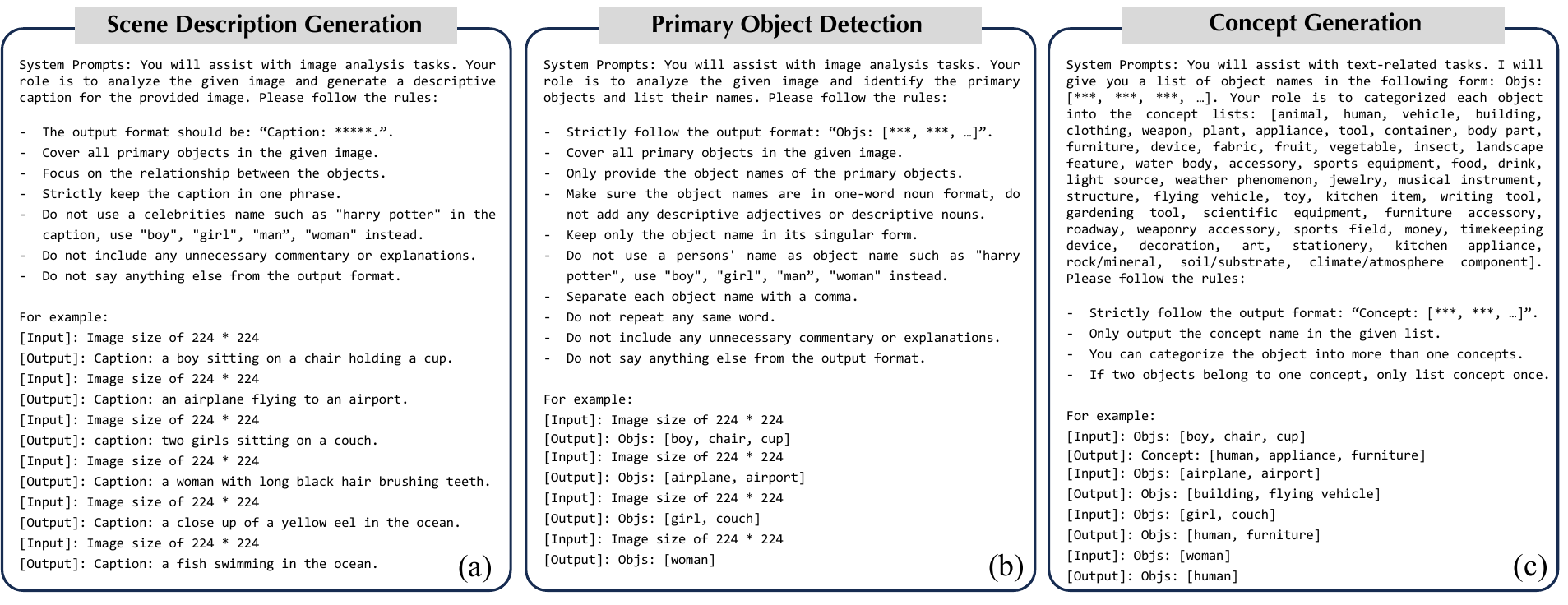}
    \caption{The overall detailed instruction prompts for scene description generation (a) and key concept generation (b-c).}
    \label{fig:prompts}
\end{figure*}

\section{Details of Instruction Prompts}\label{sec:supp_prompts}
Our \methodname{} consists of four decoupled tasks, \textit{i.e.}, scene description generation, concept name generation, segmentation mask generation, and rule-based key object discovery. However, the cc2017 dataset~\cite{wen2018neural} only contains paired fMRI and visual stimuli. To generate high-quality labels to enable the decoupled tasks, we have designed a series of detailed instruction prompts for Qwen and Grounded-SAM. The complete instruction prompts are shown in Fig.~\ref{fig:prompts}.

\section{More Details about \methodname{}}
\subsection{More Details about Brain Model}\label{sec:supp_BMdetail}

The Brain Model employs the MindEye2~\cite{scotti2024mindeye2} as the backbone, which consists of a ridge regression module, a Residual MLP module, and a diffusion prior network. 
The ridge regression module maps $x_c$ to a lower dimensional for easier follow-up, the Residual MLP module further learns the representation in a deeper hidden space, and the diffusion prior network transforms the fMRI hidden features to image embeddings.

The training of the Brain Model mainly consists of three losses: 
contrastive learning loss $\mathcal{L}_\mathrm{CLIP_t}$ between CLIP text embedding $\hat{e}^t$ and $e^t$, 
contrastive learning loss $\mathcal{L}_\mathrm{CLIP_v}$ between CLIP video embedding $\hat{e}^v\in\mathbb{R}^{BF\times N\times C}$ and $e^v$, and prior loss $\mathcal{L}_
\mathrm{prior}$. 
$\mathcal{L}_\mathrm{CLIP_t}$ and $\mathcal{L}_\mathrm{CLIP_v}$ are the implementation of BiMixCo loss which aligns all the frames of a video ${y}_c$ and its corresponding fMRI signal ${{x}_c}$ using bidirectional contrastive loss and MixCo data augmentation.
The MixCo needs to mix two independent fMRI signals.
For each ${x}_c$, we random sample another fMRI ${x}_{m_c}$, which is the keyframe of the clip index by $m_c$.
Then, we mix ${x}_c$ and ${x}_{m_c}$ using a linear combination:
\begin{equation}
    {x}^{*}_{c} = mix({x}_c, {x}_{m_c}) 
    = \lambda_c \cdot {x}_c + (1-\lambda_c) {x}_{m_c},
\end{equation}
where ${x}^{*}_{c}$ denotes mixed fMRI signal and $\lambda_c$ is a hyper-parameter sampled from Beta distribution.
Then, we adapt the ridge regression to map ${x}^{*}_{c}$ to a lower-dimensional ${x}^{*'}_{c}$ and obtain the embedding $e_{{x}^{*}_{c}}$ via the MLP, \textit{i.e.}, $e_{{x}^{*}_{c}} = \mathcal{E}({x}^{*'}_{c})$.
Based on this, the BiMixCo loss can be formed as:
\begin{footnotesize}
\begin{equation}
    \begin{aligned}
        \mathcal{L}_\mathrm{BiMixCo} =
        & - \frac{1}{2F} \sum_{i=1}^{F}
        \lambda_i \cdot \log \frac
        { \exp( sim(e_{{x}^{*}_{i}}, e_{{y}_i})/\tau) }
        { \sum_{k=1}^{F} \exp{(sim(e_{{x}^{*}_{i}}, e_{{y}_k}}) /{\tau}) } \\
        & - \frac{1}{2F} \sum_{i=1}^{F}
        (1-\lambda_i) \cdot \log \frac
        { \exp(sim(e_{{x}^{*}_{i}}, e_{{y}_{m_i}})/{\tau}) }
        { \sum_{k=1}^{F} \exp(sim(e_{{x}^{*}_{i}}, e_{{y}_k})/{\tau})} \\
        & - \frac{1}{2F} \sum_{j=1}^{F}
        \lambda_j \cdot \log \frac
        { \exp(sim(e_{{x}^{*}_{j}}, e_{{y}_j}) / {\tau}) }
        { \sum_{k=1}^{F} \exp(sim(e_{{x}^{*}_{k}}, e_{{y}_j})/{\tau}) } \\
        & - \frac{1}{2F} \sum_{j=1}^{F}
        \sum_{\{l|m_l=j\}}
        (1-\lambda_j) \\ 
        & \cdot \log \frac
        { \exp(sim(e_{{x}^{*}_{l}}, e_{{y}_j}) /{\tau}) }
        { \sum_{k=1}^{F} \exp(sim(e_{{x}^{*}_{k}}, e_{{y}_j})/{\tau})},
    \end{aligned}
\end{equation}
\end{footnotesize}
where $\hat{e}^{t}\in\mathbb{R}^{BF\times N\times C}$ denotes the OpenCLIP embeddings for video ${y}_c$.

We use the Diffusion Prior to transform fMRI embedding $e_{{x}_c}$ into the reconstructed OpenCLIP embeddings of video $e^{v}$.
Similar to DALLE·2, Diffusion Prior predicts the target embeddings with mean-squared error (MSE) as the supervised objective:
\begin{equation}
    \mathcal{L}_\mathrm{Prior} = \mathbb{E}_{e_{{y}_c}, e_{{x}_c}, \epsilon \sim \mathcal{N}(0,1)} 
    || \epsilon(e_{{x}_c}) - e_{{y}_c} ||.
\end{equation}

\subsection{Key Object Segmentation Objective} \label{sec:supp_seg}
\begin{equation}
\begin{aligned}
    \mathcal{L}_{seg}(y^{seg}, \hat{y}^{seg}) &= -\frac{1}{B\times F} \sum_{i=1}^{B \times F} \left[ y_i^{seg} \log(\hat{y}_i^{seg}) \right.\\
    &\left.+ (1 - y_i^{seg}) \log(1 - \hat{y}_i^{seg}) \right]
    \end{aligned}
\end{equation}
where $\hat{y}^{seg}$ is the ground truth masks of the key objects.

\section{Details of Data Pre-processing}\label{sec:supp_preprocessing}  
We utilized fMRI data from the cc2017 dataset, preprocessed by ~\cite{gong2025neuroclips} using the minimal preprocessing pipeline~\cite{glasser2013minimal}. The preprocessing steps included artifact removal, motion correction (6 degrees of freedom), registration to standard space (MNI space), and transformation onto cortical surfaces, which were coregistered to a cortical surface template~\cite{glasser2016multi}. To identify stimulus-activated voxels, we computed the voxel-wise correlation between fMRI signals for each repetition of the training movie across subjects. The correlation coefficients for each voxel were Fisher z-transformed, and the average z-scores across 18 training movie segments were evaluated using a one-sample t-test. Voxels with significant activation (Bonferroni-corrected, P $<$ 0.05) were selected for further analysis. This process identified 13,447, 14,828, and 9,114 activated voxels in the visual cortex for the three subjects, respectively. Consistent with prior studies~\cite{nishimoto2011reconstructing,han2019vaevideo,wang2022cgan}, we incorporated a 4-second delay in the BOLD signals to account for hemodynamic response latency when mapping movie stimulus responses.

\section{Implementation Details}\label{sec:supp_inplement} 
In this paper, videos from the cc2017 dataset were downsampled from 30FPS to 3FPS to make a fair comparison with the previous methods, and the blurred video was interpolated to 8FPS to generate the final 8FPS video during inference. 
The training of the Brain Model and Decoupler was performed with 30 and 50 epochs, respectively, and the batch size of training the Brain Model was set to 120, while 10 for the Decoupler. 
We use the AdamW~\cite{loshchilov2017adamw} for
optimization, with a learning rate set to 5e-5, to which the OneCircle learning rate schedule~\cite{smith2019onecircle} was set. 
Theoretically, our approach can be used in any text-to-video diffusion model, and we choose the open-source available AnimateDiff~\cite{guo2023animatediff} as our inference model following \cite{gong2025neuroclips}. The inference is performed with 25 DDIM~\cite{song2020ddim} steps. All experiments were conducted using a single A100 GPU.

\begin{figure}[ht]
    \centering
    \includegraphics[width=1\linewidth]{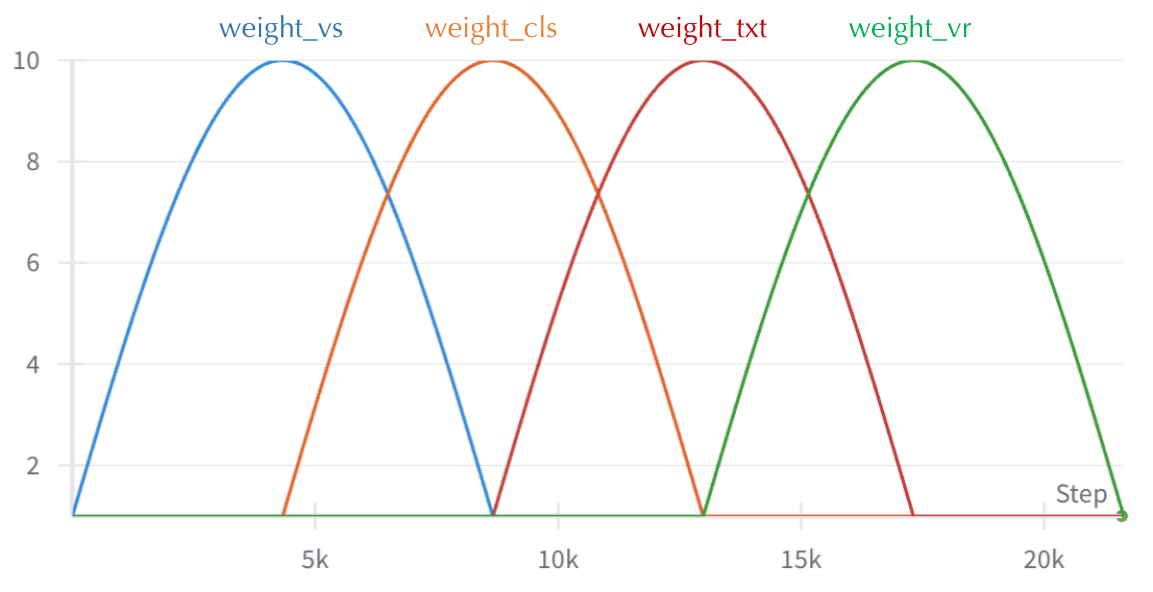}
    \caption{Illustration of the progressive learning strategy.}
    \label{fig:weights}
\end{figure}

\begin{table}[ht]
  \caption{Quantitative comparison of \methodname{} reconstruction performance against other methods (Frame-based).  Bold font signifies the best performance, while underlined text indicates the second-best performance. MinD-Video and \textit{NeuroClips} are both results averaged across all three subjects, and the other methods are results from subject 1. Results of baselines are quoted from~\cite{gong2025neuroclips}.}
\resizebox{\linewidth}{!}{
  \begin{tabular}{lcccc}
    \toprule
    \multirow{2}{*}{Method} & \multicolumn{2}{c}{Semantic-level} & \multicolumn{2}{c}{Pixel-level} \\
    \cmidrule(lr){2-3} \cmidrule(lr){4-5}
    & 2-way$\uparrow$ & 50-way$\uparrow$ & SSIM$\uparrow$ & PSNR$\uparrow$ \\
    \midrule
    Nishimoto~\cite{nishimoto2011reconstructing} & 0.727\scriptsize{$\pm0.04$} & - & 0.116\scriptsize{$\pm0.09$} & 8.012\scriptsize{$\pm2.31$} \\
    Wen~\cite{wen2018neural} & 0.758\scriptsize{$\pm0.03$} & 0.070\scriptsize{$\pm0.01$} & 0.114\scriptsize{$\pm0.15$} & 7.646\scriptsize{$\pm3.48$} \\
    Wang~\cite{wang2022reconstructing} & 0.713\scriptsize{$\pm0.04$} & - & 0.118\scriptsize{$\pm0.08$} & \textbf{11.432\scriptsize{$\pm2.42$}} \\
    Kupershmidt~\cite{kupershmidt2022penny} & 0.764\scriptsize{$\pm0.03$} & 0.179\scriptsize{$\pm0.02$} & 0.135\scriptsize{$\pm0.08$} & 8.761\scriptsize{$\pm2.22$} \\
    MinD-Video~\cite{chen2024cinematic} 
    & {0.796\scriptsize{$\pm0.03$}} & {0.174\scriptsize{$\pm0.03$}} & {0.171\scriptsize{$\pm0.08$}} & 8.662\scriptsize{$\pm1.52$} \\
    NeuroClips~\cite{gong2025neuroclips} 
    & \underline{0.806\scriptsize{$\pm0.03$}} & \underline{0.203\scriptsize{$\pm0.01$}} & \textbf{0.390\scriptsize{$\pm0.08$}} & {9.211\scriptsize{$\pm1.46$}} \\
    
    \rowcolor{cyan!10} 
    \textbf{\methodname{} (ours)} 
    & \textbf{0.811\scriptsize{$\pm0.03$}} & \textbf{0.210\scriptsize{$\pm0.01$}} & \underline{0.365\scriptsize{$\pm0.11$}} & \underline{9.527\scriptsize{$\pm2.26$}} \\
    \midrule
    \rowcolor{cyan!10} 
    subject 1 & {0.810}\scriptsize{$\pm0.03$} & {0.206}\scriptsize{$\pm0.01$} & {0.373}\scriptsize{$\pm0.14$} & {9.591}\scriptsize{$\pm2.24$} \\
    \rowcolor{cyan!10} 
    subject 2 & {0.810}\scriptsize{$\pm0.03$} & {0.214}\scriptsize{$\pm0.01$} & {0.353}\scriptsize{$\pm0.08$} & {9.502}\scriptsize{$\pm2.40$} \\
    \rowcolor{cyan!10} 
    subject 3 & {0.817}\scriptsize{$\pm0.03$} & {0.210}\scriptsize{$\pm0.01$} & {0.369}\scriptsize{$\pm0.13$} & {9.488}\scriptsize{$\pm2.16$} \\
    \bottomrule
  \end{tabular}}
\label{tab:frame}
\end{table}

\section{More Experimental Results}

\subsection{Frame-based comparison with SOTAs.}\label{sec:supp_frame_comp}
We compare the frame-based evaluation metrics in Table~\ref{tab:frame} with previous SOTA methods. \methodname{} consistently excels in semantic-level frame understanding while maintaining competitive pixel-level performance. As shown in Table 1, our method achieves the highest Semantic-level 2-way accuracy (0.811), outperforming NeuroClips by 0.6\% and MinD-Video by 1.9\%. For 50-way semantic classification, \methodname{} attains 0.210, a 3.4\% improvement over NeuroClips, highlighting its capability to discern fine-grained semantic patterns. In pixel-level metrics, \methodname{} achieves the second-best PSNR (9.527), surpassing NeuroClips by 3.4\%, while its SSIM (0.365) remains competitive. The observed trade-off between Wang’s high PSNR (11.432), and its poor semantic performance (0.713 in 2-way) underscores the challenge of balancing reconstruction fidelity with semantic alignment—a challenge \methodname{} addresses effectively through its unified architecture. These results validate that our approach advances the state of the art in frame-based video understanding by harmonizing semantic and low-level feature learning.

\subsection{More Qualitative Comparison Results}\label{sec:supp_morevis}
\begin{figure*}[ht]
    \centering
    \includegraphics[width=0.9\linewidth]{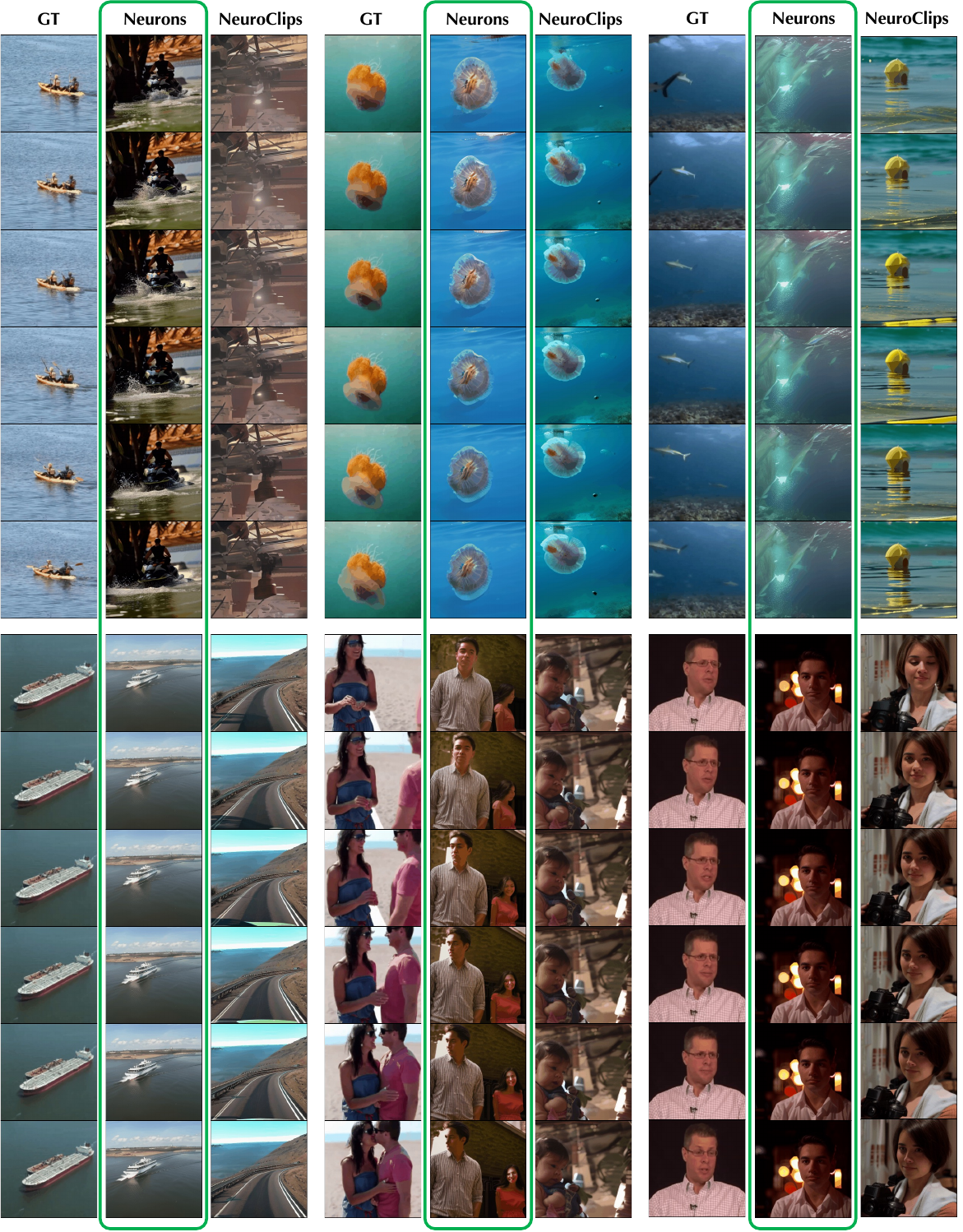}
    \caption{More qualitative comparison between \methodname{} and the previous SOTA, NeuroClips.}
    \label{fig:supp_more_vis}
\end{figure*}

To further highlight the superior performance of \methodname{}, we provide additional qualitative comparisons between our method and the previous SOTA approach, NeuroClips~\cite{gong2025neuroclips}, as a supplement to Fig.~\ref{fig:SOTA_vis}. As we can see in Fig.~\ref{fig:supp_more_vis}, NeuroClips produces many semantic errors (\textit{e.g.}, confusing a ``boat'' with a ``highway road''), whereas \methodname{} produces more accurate and visually coherent results.

\begin{figure}[tb]
    \centering
    \includegraphics[width=1\linewidth]{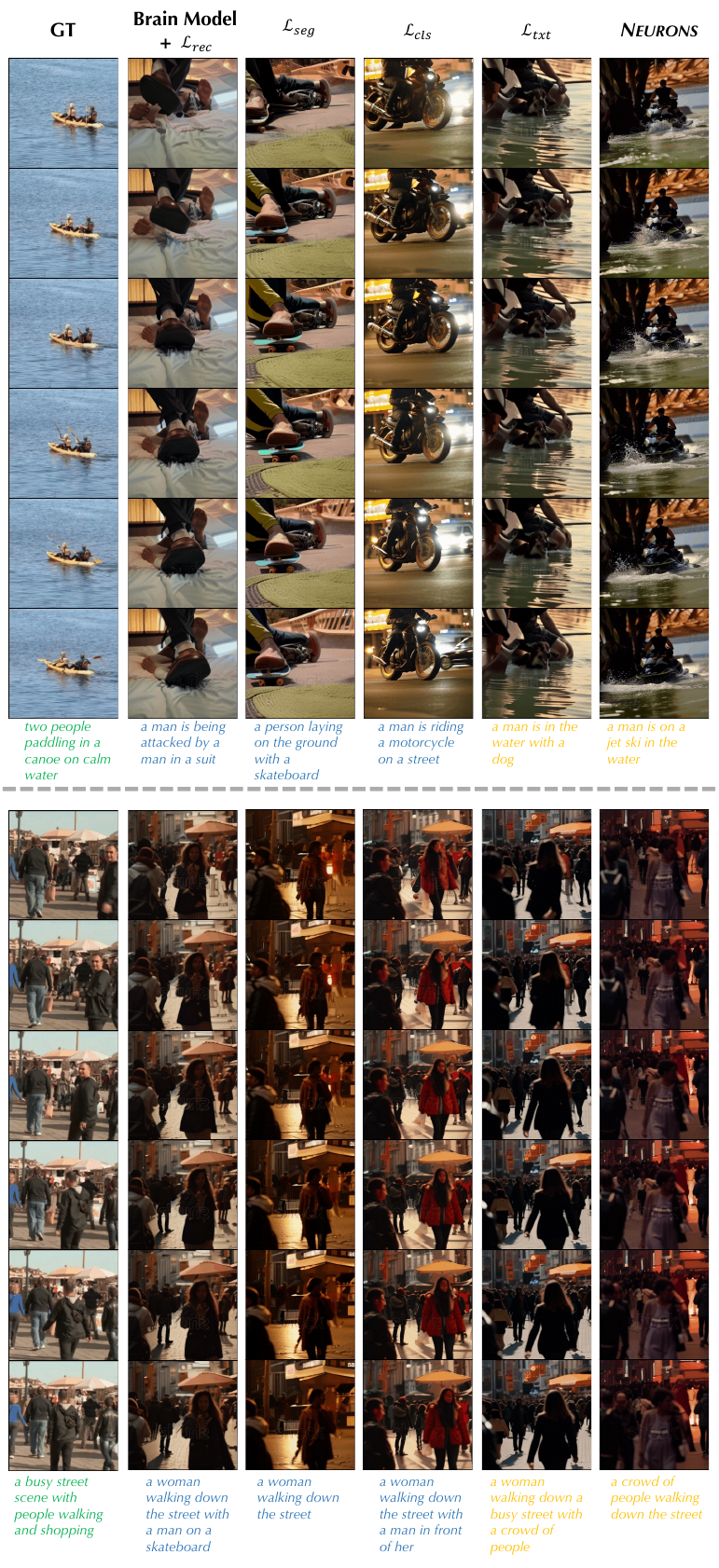}
    \caption{We present the generated videos alongside their corresponding text descriptions. Note that the descriptions in the last two columns are generated using our model $\mathcal{D}_{txt}$ (highlighted in yellow), while the other descriptions are produced by captioning the middle frame using BLIP-2.}
    \label{fig:supp_ablation}
\end{figure}

\subsection{Qualitative Analysis of Ablation Study}\label{sec:supp_ablation}
We present qualitative results from the ablation study, as illustrated in Fig.~\ref{fig:supp_ablation}. 
From the top case, the results demonstrate that the Brain Model utilizing only $\mathcal{L}_{rec}$ exhibits limited semantic information. The inclusion of $\mathcal{L}_{seg}$ could capture more motion information. 
$\mathcal{L}_{cls}$ improves the accuracy of concept recognition, such as distinguishing between humans and vehicles. Further incorporating $\mathcal{L}_{txt}$ enables the model to perceive broader scenes and environments, such as bodies of water. Finally, the combination of progressive learning and the aggregated video reconstruction pipeline ensures that the video output is both semantically and spatially accurate.
In the more complex scene from the bottom case, where the main elements (humans, streets) are easier to identify, the reconstructed videos show no significant visual differences. However, we observe that $\mathcal{L}_{seg}$ tends to make the model concentrate on key objects (\textit{e.g.}, only one person generated in the video). In contrast, the loss functions $\mathcal{L}_{cls}$ and $\mathcal{L}_{txt}$ contribute to enriching the richness of concepts and details within the scene. We also provide the corresponding descriptions for the videos. It is evident that the descriptions generated by our trained GPT-2 model are more accurate. For example, in the bottom case, it successfully generates terms like ``people'' and ``busy street'' which are consistent with the ground truths.

\subsection{Concept Recognition Accuracy.}\label{sec:supp_cls}
For one of our decoupled tasks, \textit{i.e.}, concept recognition, we provide the accuracy score for each concept as a supplement to Table.~\ref{tab:combined_results}. The results are shown in Table.~\ref{tab:all_class_accuracy}.

\subsection{Details of Caption and Verb Evaluation}\label{sec:supp_capverb_detail}
For caption evaluation, we follow traditional metrics for image captioning~\cite{chen2015microsoft}, and report BLEU and CIDEr scores. BLEU analyzes the co-occurrences of n-grams between the candidate and reference sentences, and CIDEr employs TF-IDF weighting to focus more on semantically informative words that capture image-specific contents.

To better analyze the model's motion understanding capabilities, we specifically evaluate verb accuracy within the generated captions, providing a deeper insight into how effectively the model identifies dynamic actions during video reconstruction. We first extract verbs from both generated and ground truth captions using the part-of-speech (POS) tagging model and then adopt Word2Vec embedding to calculate semantic similarity between the verb pairs. The generated verb is considered correct if the similarity score exceeds the pre-defined threshold of 0.8.

\section{Details of Brain Decoding Interpretation} \label{sec:Interpretation}
To validate the inspiration behind our \methodname{}, which is drawn from the human visual cortex (see Fig.\ref{fig:abstract} in the Introduction section), we employ a visualization tool, \textit{i.e.}, BrainDecodesDeepNets~\cite{yang2024braindecodesdeepnets}, to project the embeddings of each decoupled task onto a brain map. Specifically, we use fMRI data as input and extract four projected embeddings corresponding to the four decoupled tasks of \methodname{}. Next, we train a BrainNet~\cite{yang2024braindecodesdeepnets} to reconstruct the original fMRI image. Finally, following the approach of BrainDecodesDeepNets, we visualize the individual layer weights for each decoupled task (see Fig.~\ref{fig:interpretation}), further confirming our initial insights. For the training of the brainNet, we utilize the Algonauts 2023 challenge dataset, the same as BrainDecodesDeepNets~\cite{yang2024braindecodesdeepnets}. We train it for 50 epochs using AdamW optimizer on 1 RTX3090 GPU card. The learning rate is set to 1e-5.

\section{Limitations and Future Work} 
While \methodname{} demonstrates strong performance in fMRI-to-video reconstruction, several limitations remain. First, the model is evaluated on a single dataset with limited subject diversity, restricting generalizability. Second, the reconstructed videos, though temporally smooth, remain low in resolution and frame rate due to the coarse temporal granularity of fMRI data. Additionally, current evaluation metrics may not fully capture perceptual realism or narrative coherence. Lastly, although the model shows functional alignment with visual cortex regions, the neurobiological interpretation remains correlational rather than causal.

\begin{table}[ht]
\centering
\caption{Classification accuracy of all the concepts. "-" denotes no such concept in the test set.}
\resizebox{!}{0.6\textwidth}{
\begin{tabular}{c|c|c}
\toprule
\textbf{Index} & \textbf{Class Name} & \textbf{Accuracy} \\ \midrule
0 & animal & 0.450 \\ 
1 & human & 0.735 \\ 
2 & vehicle & 0.228 \\ 
3 & building & 0.070 \\ 
4 & clothing & 0.184 \\ 
5 & weapon & - \\ 
6 & plant & 0.196 \\ 
7 & appliance & - \\ 
8 & tool & 0.0625 \\ 
9 & container & 0.057 \\ 
10 & body part & 0.163 \\ 
11 & furniture & 0.155 \\ 
12 & device & 0.033 \\ 
13 & fabric & - \\ 
14 & fruit & 0.0 \\ 
15 & vegetable & 0.0 \\ 
16 & insect & - \\ 
17 & landscape feature & 0.144 \\ 
18 & water body & 0.310 \\ 
19 & organism & 0.212 \\ 
20 & fish & 0.292 \\ 
21 & reptile & 0.0623 \\ 
22 & mammal & - \\ 
23 & accessory & 0.067 \\ 
24 & sports equipment & 0.182 \\ 
25 & food & 0.6 \\ 
26 & drink & 0.0 \\ 
27 & light source & 0.0 \\ 
28 & weather phenomenon & 0.091 \\ 
29 & jewelry & - \\ 
30 & musical instrument & - \\ 
31 & structure & 0.209 \\ 
32 & flying vehicle & 0.283 \\ 
33 & toy & - \\ 
34 & kitchen item & 0.214 \\ 
35 & writing tool & - \\ 
36 & gardening tool & - \\ 
37 & scientific equipment & 0.0 \\ 
38 & furniture accessory & 0.0 \\ 
39 & roadway & 0.147 \\ 
40 & weaponry accessory & - \\ 
41 & sports field & 0.042 \\ 
42 & money & - \\ 
43 & timekeeping device & - \\ 
44 & decoration & - \\ 
45 & art & 0.0 \\ 
46 & stationery & 0.111 \\ 
47 & kitchen appliance & - \\ 
48 & rock/mineral & 0.0 \\ 
49 & soil/substrate & 0.0 \\ 
50 & climate/atmosphere component & 0.262 \\ \bottomrule
\end{tabular}}
\label{tab:all_class_accuracy}
\end{table}

\clearpage
\section{Overview of AI Methodologies for Neuroscience Readers} 
This study leverages several state-of-the-art AI models to decode and reconstruct visual experiences from brain activity. At the core is a neural network-based brain model trained using \textit{contrastive learning}, which aligns fMRI signals with visual and textual embeddings from \textit{CLIP}, a widely used vision-language model. The decoding process is decomposed into four explicit sub-tasks—segmentation, classification, captioning, and reconstruction—each modeled by deep learning modules optimized with task-specific loss functions. For video synthesis, we use \textit{diffusion models}, a class of generative models that produce high-quality video frames conditioned on the outputs of the brain model. These AI components are organized in a biologically inspired, hierarchical manner to simulate the functional specialization of the human visual cortex.